\pdfoutput=1
\documentclass[10pt, letterpaper, twocolumn]{article}

\usepackage{placeins}
\usepackage{amsmath}
\usepackage{tabularx}
\usepackage{booktabs}
\usepackage{url}
\usepackage{graphicx}
\usepackage{caption}
\usepackage{subcaption}
\usepackage{xspace}
\usepackage{enumitem}
\usepackage{xcolor}
\usepackage{algorithm}
\usepackage{algorithmic}
\usepackage{hyperref} 
\usepackage{amssymb}

\usepackage[acronym,nogroupskip,nonumberlist,nopostdot]{glossaries}

\newacronym{DRL}{DRL}{deep reinforcement learning}
\newacronym{MARL}{MARL}{multi-agent reinforcement learning}
\newacronym{MAPS}{MAPS}{master-agent proto-plan system}

\begin{document}

\title{Compact Latent Coordination for Autonomous Vehicles at Unsignalized Intersections}

\author{
    Gil Lifshits \\
    Ben-Gurion University of the Negev \\
    \texttt{gillif@post.bgu.ac.il}
    \and
    Igal Bilik \\
    Ben-Gurion University of the Negev \\
    \texttt{bilik@bgu.ac.il}
    \and
    Gilad Katz \\
    Ben-Gurion University of the Negev \\
    \texttt{giladkz@bgu.ac.il}
}

\maketitle 

\begin{abstract}
Coordinating autonomous vehicles at unsignalized intersections remains a critical challenge for \gls{MARL} systems, which typically struggle with combinatorial action spaces, reliance on privileged information, or rigid agent designs. We propose Master-Agent Proto-plan System (MAPS), a hierarchical \gls{DRL} architecture in which a centralized Master agent generates a compact, continuous embedding, denoted as \emph{proto-plan}, that encodes a global coordination strategy. Decentralized Worker agents integrate this embedding with local observations to execute vehicle-specific control, decoupling strategic intent from tactical execution and enabling independent optimization of each module.
As a proof-of-concept evaluation of this coordination mechanism, we test MAPS across 72 intersection configurations in \textit{HighwayEnv}. MAPS achieves collision-free navigation while significantly reducing average travel time, outperforming state-of-the-art baselines. The learned \emph{proto-plans} further exhibit robust generalization: a system trained with three agents achieves a 94\% success rate when deployed zero-shot to five-agent scenarios, confirming that \emph{proto-plan}-based hierarchical learning provides a promising framework for multi-vehicle coordination.
\end{abstract}

\section{Introduction}

Coordinating multiple autonomous vehicles (AVs) at unsignalized intersections is a central challenge for intelligent transportation: vehicles must jointly navigate complex, dynamic scenarios while guaranteeing collision-free operation and maintaining throughput~\cite{Wang2024}. The difficulty of this multi-agent coordination problem has driven extensive research in multi-agent deep reinforcement learning (MADRL)~\cite{Chen2024}, spanning value-decomposition methods~\cite{Huang2023Multiagent}, graph-based representations~\cite{Cai2022DQGATTowardS}, policy optimization~\cite{Peng2023Curriculum, Xu2022DecisionMaking}, and hierarchical frameworks~\cite{AlSharman2022SelfLearned, Liu2025Cooperative, Zhao2024CentralizedCF}.

Despite notable progress, existing approaches share several limitations. First, many exhibit \emph{limited generalization}, as they are trained and evaluated on simplified layouts that do not reflect the real-world diversity of intersections. Second, state-of-the-art methods often depend on \emph{privileged information}, such as future trajectories, game-theoretic priors, rule-based safety layers, or expert demonstrations, that may be unavailable in practical deployments. Third, most methods employ \emph{fixed, discrete action spaces} consisting of small sets of predefined maneuvers, which scale combinatorially with the number of agents and require redesign when vehicle capabilities change.

We address these limitations with \gls{MAPS}, a hierarchical \gls{DRL} architecture that replaces explicit, per-vehicle action assignments with \emph{proto-plans}: learned, continuous embedding vectors that encode high-level coordination strategies. A centralized Master agent observes the global traffic state and produces a single \emph{proto-plan}; decentralized Worker agents each combine this \emph{proto-plan} with their own local observations to select vehicle-specific actions. This decomposition decouples coordination from control, keeps communication overhead constant at $O(d)$ regardless of fleet size, and allows each module to be updated independently.

Experiments across 72 intersection configurations show that MAPS achieves zero collisions during evaluation while reducing average travel time to 7.8 steps, which is a 38\% improvement over the best baseline. The architecture further demonstrates robust zero-shot transfer: trained with only three active agents, it attains a 94\% success rate when deployed to five-agent scenarios without fine-tuning.

\noindent\textbf{Our contributions are as follows:}
\begin{itemize}[nosep,leftmargin=*]
    \item We introduce MAPS, a hierarchical \gls{DRL} architecture that replaces rigid coordination commands with continuous \emph{proto-plan} embeddings, keeping action and communication complexity at $O(d)$ as fleet size grows.
    \item We demonstrate zero-shot generalization through incremental training: a model trained on three agents transfers directly to five-agent deployment, confirming that the learned \emph{proto-plans} capture transferable coordination strategies.
    \item We show that effective multi-vehicle coordination requires only readily available kinematic state (positions and velocities), with no dependence on privileged information or expert demonstrations.
\end{itemize}

\noindent\textbf{Scope.} This work introduces the \emph{proto-plan} coordination mechanism as a foundational advancement in hierarchical multi-agent systems. Our primary contribution is the architectural demonstration that continuous, learned embeddings provide a more robust and scalable coordination interface than traditional discrete command structures. By isolating the multi-agent coordination challenge from the complexities of raw perception and low-level vehicle dynamics, we provide a rigorous validation of the \emph{proto-plan’s} efficacy under controlled simulation conditions. This design choice allows us to confirm that the observed performance gains, including collision-free navigation and zero-shot transfer, are intrinsic to the hierarchical architecture itself. While this work serves as a definitive proof-of-concept for the MAPS framework, the architecture is designed for modular extensibility to high-fidelity environments and complex sensor suites, as discussed in Section~\ref{subsec:limitations}.

\section{Related Work}
\subsection{Multi-Agent Reinforcement Learning}
\gls{MARL} extends single-agent RL to settings with multiple decision-makers facing partial observability, non-stationary dynamics, and cooperation--competition tradeoffs. Recent work targets scalability, safety, and coordination in large-scale systems~\cite{Liuetal2024}.
Model-based approaches include Ma et al.~\cite{ma2024network_control}, who topologically decouple global dynamics for local model learning while approximating global value information, improving sample efficiency at scale. For safety-critical settings, Zhang et al.~\cite{zhang2024scpo} enforce joint constraints via local trust-region updates over $\kappa$-hop neighborhoods, enabling decentralized training without centralized critics, while Hsu and Pajic~\cite{hsu2025safe_marl} provide regret guarantees for safe cooperative \gls{MARL} with function approximation.

Graph-based representations encode structural priors effectively. GNN-based methods support resilient multi-robot coordination under agent failures and communication disturbances~\cite{weil2024graph_marl}. In traffic management, decentralized graph-based \gls{MARL} with traffic digital twins improves signal timing in large networks~\cite{wang2025hypergraph_tsc}, and EECG~\cite{peng2025eecg} integrates GNNs with curiosity-driven exploration and evolutionary optimization for improved credit assignment in partially observable tasks.
These approaches address important coordination aspects but typically operate within fixed action-space structures requiring agents to reason over peer-scaling representations. Our work is complementary: rather than improving the learning algorithm, we introduce a hierarchical communication mechanism via the \emph{proto-plan} embedding, compressing coordination into a constant-size signal that allows more effective application of existing RL algorithms.

\subsection{MARL for Autonomous Driving}
Applying \gls{MARL} to autonomous driving demands safety, efficiency, and real-time responsiveness. Xu et al.~\cite{Xu2022DecisionMaking} propose DDPG-based algorithms with meta-exploration and twin-delayed variants for intersection navigation, highlighting reward engineering for multi-objective optimization. In value decomposition, QMIXwD~\cite{Huang2023Multiagent} integrates self-generated demonstrations for improved early exploration, while VN-MADDPG~\cite{Zhang2024VNMADDPGAV} extends MADDPG with variable-noise and importance-sampling for more efficient multi-vehicle learning.
Graph-based models effectively capture vehicle interactions. DQ-GAT~\cite{Cai2022DQGATTowardS} leverages bird's-eye-view maps and graph attention for complex spatial relationships, and Spatharis and Blekas~\cite{Spatharis2024Multiagent} propose collaborative frameworks with route agents and collision terms for scalable SUMO coordination. For policy optimization, Peng et al.~\cite{Peng2023Curriculum} introduce curriculum PPO with stage-decaying clipping across difficulty levels, and Xu et al.~\cite{Xu2022DecisionMaking} combine TD3 with LSTM-based motion prediction for smoother trajectories.

Hierarchical architectures, particularly relevant to our work, decouple strategic from tactical decision-making. Al-Sharman et al.~\cite{AlSharman2022SelfLearned} separate high-level behavioral planning from low-level control via SAC-MPC hierarchy. MA-GA-DDPG~\cite{Liu2025Cooperative} augments MADDPG with multi-head attention and level-$k$ game priors, incorporating a safety inspector for cooperative CAV behavior. SafeR-ADAIM~\cite{Zhao2024CentralizedCF} demonstrates risk-aware constrained optimization for dense intersection safety.
These hierarchical methods typically issue \emph{discrete} commands or explicit sub-goals, yielding action spaces that grow combinatorially with fleet size and require redesign for new behaviors. Similarly, peer-to-peer learned communication methods like CommNet~\cite{commNet} and TarMAC~\cite{TarMAC} require each agent to send and receive messages, with per-agent aggregation scaling with peer count. MAPS departs from both paradigms: the Master produces a \emph{continuous} \emph{proto-plan} embedding encoding coordination intent in fixed-dimensional latent space. This avoids combinatorial scaling, requires no predefined command vocabulary, keeps Worker input size constant regardless of fleet size, and, as shown in Section~\ref{sec:main_results}, enables zero-shot transfer to larger agent populations.

\section{Proposed Approach}
\label{sec:method}

\subsection{Overview}

\begin{figure}[t]
    \centering
    \includegraphics[width=0.95\linewidth]{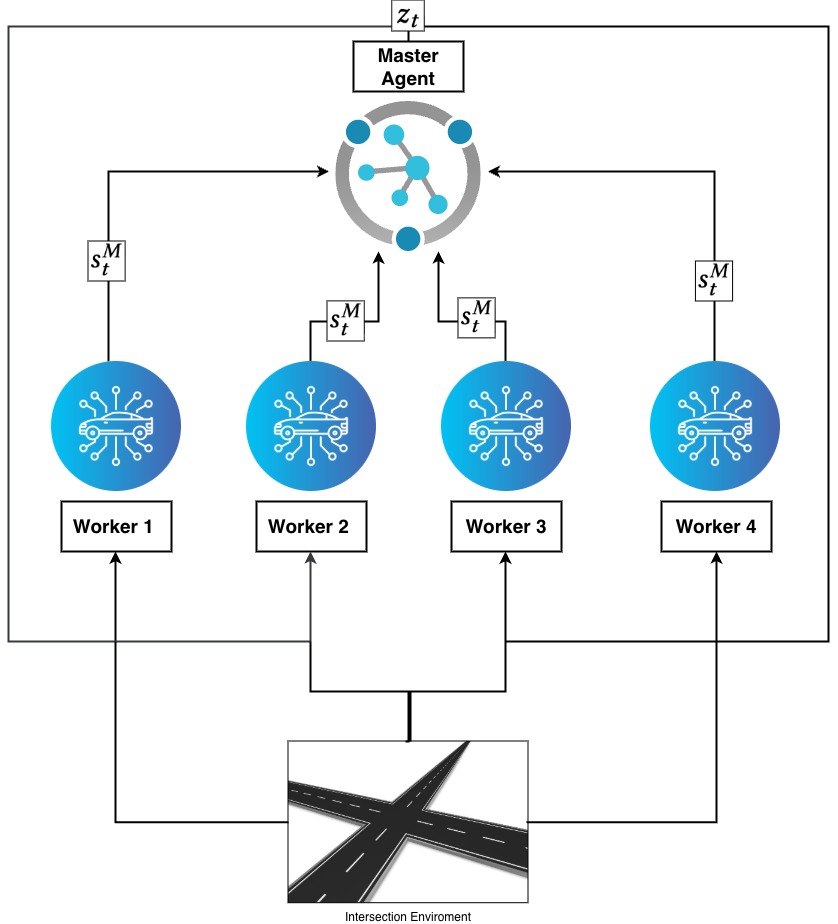}
    \caption{Hierarchical MARL framework. The Master agent observes global state $s_t^M$ and generates a \emph{proto-plan} embedding $z_t$. Each Worker $i$ receives $z_t$ along with its local observation to produce vehicle-specific actions $a_t^i$.}
    \label{fig:hierarchical_MARL_framework_figure}
\end{figure}

We formulate multi-vehicle intersection coordination as a hierarchical Markov decision process comprising two levels (Figure~\ref{fig:hierarchical_MARL_framework_figure}): a \textbf{centralized Master agent} that observes the global traffic state and generates a \emph{proto-plan} embedding $z_t \in \mathbb{R}^d$ representing the desired coordination strategy, and \textbf{decentralized Worker agents}, each controlling an individual vehicle, that combine the \emph{proto-plan} with local observations to select vehicle-specific actions. The Master's action space is a continuous latent space rather than an enumeration of per-vehicle commands, avoiding the $|\mathcal{A}|^N$ combinatorial scaling of joint action spaces while keeping communication overhead at $O(d)$ regardless of fleet size.

\subsection{Problem Formulation}
\label{subsec:problem_formulation}

We model the problem as a decentralized partially observable Markov decision process (Dec-POMDP) with hierarchical control, defined by the tuple $\langle \mathcal{N}, \mathcal{S}, \{\mathcal{O}_i\}, \{\mathcal{A}_i\}, \mathcal{T}, \{R_i\}, \gamma \rangle$, where $\mathcal{N} = \{1, \ldots, N\}$ is the set of vehicles, $\mathcal{S}$ is the global state space (positions and velocities), $\mathcal{O}_i$ and $\mathcal{A}_i$ are the local observation and action spaces for Worker $i$, $\mathcal{T}: \mathcal{S} \times \mathcal{A}_1 \times \cdots \times \mathcal{A}_N \rightarrow \Delta(\mathcal{S})$ is the transition function, $R_i: \mathcal{S} \times \mathcal{A}_i \rightarrow \mathbb{R}$ is the reward for agent $i$, and $\gamma \in [0,1)$ is the discount factor.

The objective is to find policies $\{\pi_i\}_{i=1}^N$ maximizing the expected cumulative discounted reward:
\begin{equation}
    J = \mathbb{E}\left[ \sum_{t=0}^{T} \gamma^t \sum_{i=1}^{N} R_i(s_t, a_t^i) \right],
\end{equation}
subject to each Worker $i$ accessing only its local observation $o_t^i \in \mathcal{O}_i$ rather than the full state $s_t \in \mathcal{S}$. The Master bridges this information gap by observing the global state and communicating coordination intent through the \emph{proto-plan}, realizing a centralized-training-with-decentralized-execution (CTDE) paradigm~\cite{lowe2017multi, wang2025facilitating}.

\subsection{State Space Representation}
\label{subsec:state_space}

\subsubsection{Master State Space}
The Master observes a global state $s_t^M$ formed by concatenating the kinematic information of all $N$ vehicles:
\begin{equation}
    s_t^M = \left[ x_1, y_1, v_{x_1}, v_{y_1}, \ldots, x_N, y_N, v_{x_N}, v_{y_N} \right] \in \mathbb{R}^{4N},
    \label{eq:master_state}
\end{equation}
where $(x_i, y_i)$ and $(v_{x_i}, v_{y_i})$ denote the position and velocity of vehicle $i$ in a global frame centered at the intersection. Vehicles are ordered by approach direction (N, E, S, W) and then by descending proximity to the intersection center.

The input layer accommodates a fixed maximum of $N_{\max}$ vehicles. When only $k < N_{\max}$ vehicles are active, the remaining slots are zero-padded:
\begin{equation}
    s_t^M = \left[ x_1, y_1, v_{x_1}, v_{y_1}, \ldots, x_k, y_k, v_{x_k}, v_{y_k}, \mathbf{0}_{4(N_{\max}-k)} \right].
\end{equation}

\subsubsection{Worker State Space}
Each Worker $i$ observes a local state combining its own kinematics with the \emph{proto-plan}:
\begin{equation}
    s_t^{W_i} = \left[ x_i, y_i, v_{x_i}, v_{y_i}, z_t \right] \in \mathbb{R}^{4+d},
    \label{eq:worker_state}
\end{equation}
where $z_t \in \mathbb{R}^d$ is the \emph{proto-plan} embedding ($d=4$ in our experiments). Workers receive no direct observations of other vehicles; they rely entirely on the \emph{proto-plan} for coordination information. This keeps each Worker's input size constant at $4+d$ regardless of fleet size, improving both scalability and privacy.

\subsection{Action Space and Proto-Plan Mechanism}
\label{subsec:action_space}

\subsubsection{Master Action Space}
The Master's action space is defined as $\mathcal{A}_M = \mathbb{R}^d$. Given the global state $s_t^M$, the Master policy outputs a \emph{proto-plan} embedding:
\begin{equation}
    z_t = \pi_M^{\theta_M}(s_t^M),
    \label{eq:master_policy}
\end{equation}
where $\pi_M^{\theta_M}$ is parameterized by neural network weights $\theta_M$, with a tanh output layer bounding $z_t \in (-1,1)^d$.

Unlike conventional hierarchical RL that issues discrete high-level commands such as ``vehicle~1 yields, vehicle~2 proceeds'', the \emph{proto-plan} encodes coordination strategies as dense vectors in a continuous space. The semantics of individual dimensions emerge from end-to-end training rather than manual specification.

While this mechanism shares surface similarity with learned communication protocols~\cite{commNet, TarMAC}, the design differs fundamentally. First, \emph{proto-plan} communication is \emph{asymmetric and one-to-many}: a single Master broadcasts a fixed-size vector to all Workers, whereas CommNet and TarMAC employ symmetric peer-to-peer messaging with message complexity scaling with fleet size. Second, the \emph{proto-plan} is not a message \emph{about} any individual agent; it is a compressed \emph{global coordination strategy}, conceptually closer to a learned option or subgoal in the hierarchical RL sense. Third, Workers are fully decoupled from fleet composition: they receive the same $d$-dimensional input regardless of active agents, whereas peer-to-peer protocols require each agent to aggregate messages from a variable number of peers. This architectural separation means the Master absorbs all coordination complexity, allowing Worker policies to remain fleet-size-agnostic and enabling the zero-shot transfer demonstrated in Section~\ref{subsec:generalization}.

\subsubsection{Worker Action Space}
Each Worker operates with a discrete action space $\mathcal{A}_W = \{\texttt{accelerate}, \texttt{decelerate}\}$, where each action modifies the target speed by $\Delta v_{\text{target}} = \pm 5 \, m/s$. A low-level controller then computes the physical acceleration required to reach the new target speed. The control frequency is $1\,\text{Hz}$; no ``maintain speed'' option is provided.

At each time step, the Worker policy selects an action based on its local state:
\begin{equation}
    a_t^i = \pi_W^{\theta_W}(s_t^{W_i}).
    \label{eq:worker_policy}
\end{equation}

\subsection{Reward Structure}
\label{subsec:reward}

\subsubsection{Worker Reward Function}
Each Worker receives a reward encouraging safe, efficient traversal:
\begin{equation}
r_t^{W_i} =
\begin{cases}
+R_{\text{success}} & \text{if vehicle } i \text{ crosses successfully} \\
-R_{\text{collision}} & \text{if vehicle } i \text{ is involved in a collision} \\
-R_{\text{step}} & \text{otherwise (per time step)}
\end{cases}
\label{eq:worker_reward}
\end{equation}
with $R_{\text{success}} = 50$, $R_{\text{collision}} = 300$, and $R_{\text{step}} = 5$. The asymmetric magnitudes ($R_{\text{collision}} \gg R_{\text{success}} > R_{\text{step}}$) enforce a strict priority hierarchy: safety first, then efficiency, then throughput~\cite{Hua2025, Zheng2025SafeMARL}. We forego reward normalization to maintain a steep value gradient, ensuring the collision penalty strictly dominates cumulative step costs.

\subsubsection{Master Reward Function}
The Master's reward aggregates individual Worker rewards via a minimum operator:
\begin{equation}
    R_t^M = 
    \begin{cases}
    \min_{i \in \mathcal{W}_{\text{active}}(t)} r_t^{W_i} & \text{if } \mathcal{W}_{\text{active}}(t) \neq \emptyset \\
    0 & \text{otherwise}
    \end{cases}
    \label{eq:master_reward}
\end{equation}
where $\mathcal{W}_{\text{active}}(t) = \{ i \in \mathcal{N} \mid f_i(t) = 0 \}$ is the set of vehicles that have not yet reached their destination. This maximin objective forces the Master to maximize the worst-case individual outcome, preventing coordination strategies that sacrifice any single vehicle. Replacing min with mean aggregation reduces SR to 84\% during training and 80\% during evaluation.

\subsubsection{Episode Termination}
An episode terminates when (i) all vehicles reach their destinations, (ii) a collision occurs, or (iii) the 50-step limit is reached.

\subsection{Network Architecture}
\label{subsec:architecture}

Both Master and Worker agents are trained using Proximal Policy Optimization (PPO)~\cite{schulman2017proximal}. Each Worker module consists of separate policy and value networks with four fully-connected layers (64, 32, 16, 8 neurons; ReLU activations). The policy network outputs a categorical distribution over the two actions via softmax; the value network outputs a scalar estimate. The Master uses three-layer networks (128, 256, 128 neurons) to accommodate the higher-dimensional global input ($4N_{\max}$ features), with a tanh output layer producing the $d$-dimensional \emph{proto-plan}.

All Worker agents share parameters $\theta_W$, reducing learnable parameters and encouraging generalizable control strategies. Since Workers operate on local, ego-centric observations, a single policy can map diverse situational contexts to unified driving behaviors.

\subsection{Training Process}
\label{subsec:training}

Training proceeds through alternating optimization of the Master and Worker modules, as formalized in Algorithm~\ref{alg:joint-training}. In the initial cycle (cycle~0), both policies are updated jointly to establish preliminary coordination. Subsequent cycles alternate between freezing the Master (Workers adapt to current \emph{proto-plan} representations) and freezing the Workers (the Master learns to generate more effective \emph{proto-plans} given Workers' current behavior). This reduces the non-stationarity inherent in multi-agent optimization: each module trains against a fixed counterpart, yielding more reliable gradient estimates and stable convergence.

Experience is collected into a dedicated rollout buffer for the Master and individual buffers for each Worker. The complete set of training hyperparameters is provided in Table~\ref{tab:hyperparameters}.

\subsubsection{Incremental Training for Generalization}
\label{subsec:incremental}

To evaluate the transferability of learned \emph{proto-plans}, we employ an incremental training protocol. The Master's input layer is sized for $N_{\max} = 5$ agents from the outset, with unused slots zero-padded. Training proceeds in two phases:
\begin{enumerate}[nosep]
    \item \textbf{Phase 1:} Train with 1 learning agent until convergence.
    \item \textbf{Phase 2:} Introduce 2 additional agents (3 total) and continue training.
\end{enumerate}
The resulting model is then evaluated on 5-agent configurations without fine-tuning; results are reported in Section~\ref{sec:evaluation}.

\begin{algorithm}[h]
\caption{Alternating Master-Worker Training}
\label{alg:joint-training}
\textbf{Input}: Configuration $\mathcal{E}$, environment $E$, Master policy $\pi_M$, Worker policy $\pi_W$\\
\textbf{Output}: Trained policies $\pi_M$, $\pi_W$; collision count $c$; training statistics
\noindent\rule{\linewidth}{0.1pt}
\begin{algorithmic}[1]
\STATE $N \gets$ number of vehicles in $E$
\STATE Initialize rollout buffers $\mathcal{B}_M$ for Master and $\mathcal{B}_W$ for Workers
\STATE Initialize collision count $c \gets 0$
\STATE Initialize statistics container $\mathcal{S}$
\FOR{$\text{cycle} = 0$ \TO $\mathcal{E}.\text{NUM\_CYCLES} - 1$}
    \FOR{$\text{episode} = 1$ \TO $\mathcal{E}.\text{EPISODES\_PER\_CYCLE}$}
        \STATE $(r_{\text{ep}}, \tau) \gets \textsc{ExecuteEpisode}(\pi_M, \pi_W, E)$
        \STATE Store trajectory $\tau$ in appropriate buffers
        \IF{collision detected in $\tau$}
            \STATE $c \gets c + 1$
        \ENDIF
        \STATE Record episode statistics in $\mathcal{S}$
        
        \IF{$\text{cycle} = 0$}
            \STATE \textsc{UpdatePPO}($\pi_M, \mathcal{B}_M$); \textsc{UpdatePPO}($\pi_W, \mathcal{B}_W$) \hfill \COMMENT{Joint}
        \ELSIF{$\text{cycle} \bmod 2 = 0$}
            \STATE \textsc{UpdatePPO}($\pi_M, \mathcal{B}_M$) \hfill \COMMENT{Master only}
        \ELSE
            \STATE \textsc{UpdatePPO}($\pi_W, \mathcal{B}_W$) \hfill \COMMENT{Workers only}
        \ENDIF
        
        \STATE $E.\textsc{Reset}()$
    \ENDFOR
\ENDFOR
\STATE \textbf{return} $\pi_M$, $\pi_W$, $c$, $\mathcal{S}$
\end{algorithmic}
\end{algorithm}

\section{Experimental Setup}
\label{sec:experimental_setup}

\subsection{Simulation Environment}
\label{subsec:simulation_env}

We evaluate our approach using the \emph{HighwayEnv} simulator~\cite{highway-env}, an open-source reinforcement learning environment for autonomous driving research. Its modular architecture permits customization of state representations, action spaces, and reward functions, making it well-suited to our hierarchical framework. We deliberately employ \emph{HighwayEnv}'s simplified kinematics as a feature: by abstracting away perception pipelines and detailed vehicle dynamics, the simulator isolates the multi-agent coordination challenge, providing a controlled testbed where performance differences can be attributed to coordination architecture rather than confounding factors.

\subsection{Scenario Configuration}
\label{subsec:scenario_config}

\subsubsection{Intersection Scenarios}
We constructed 18 base scenarios varying in approach direction (N, S, E, W), initial distance from the intersection, and turning intention (straight, left, right). Each base scenario is rotated through all four cardinal orientations (0\textdegree, 90\textdegree, 180\textdegree, 270\textdegree), yielding \textbf{72 unique configurations}. This rotation augmentation prevents overfitting to specific approach directions and ensures generalization across symmetric traffic patterns.

\subsubsection{Scenario Design Principles}
Scenarios were hand-designed to satisfy two properties. First, \emph{solvability}: each scenario admits at least one collision-free coordination strategy, validated by executing rule-based policies. Second, \emph{non-triviality}: all scenarios feature inherently conflicting trajectories that cannot be resolved by simple reactive behaviors such as fixed yielding. The high collision rates of both baselines on these scenarios (Table~\ref{tab:crash_comparison}) empirically confirm that non-trivial coordination is required.

\subsubsection{Vehicle Initialization and Agent Assignment}
\begin{figure}[h]
    \centering
    \begin{subfigure}[t]{0.48\linewidth}
        \centering
        \includegraphics[width=\linewidth,height=4.2cm]{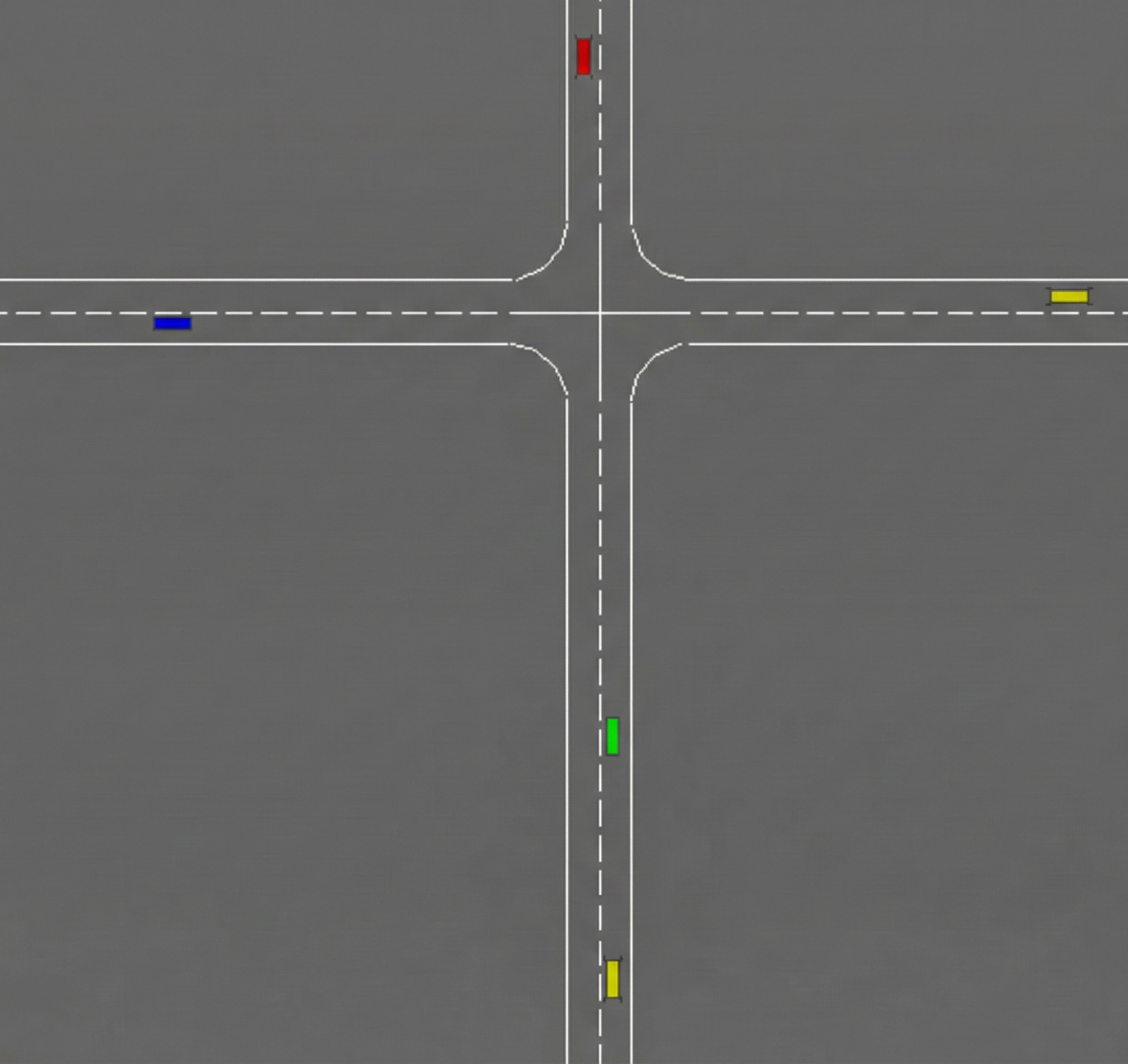}
        \caption{
         Initial configuration: five vehicles approach from all directions with varying distances to the intersection.
        }
        \label{fig:scenario_a}
    \end{subfigure}
    \hfill
    \begin{subfigure}[t]{0.48\linewidth}
        \centering
        \includegraphics[width=\linewidth,height=4.2cm]{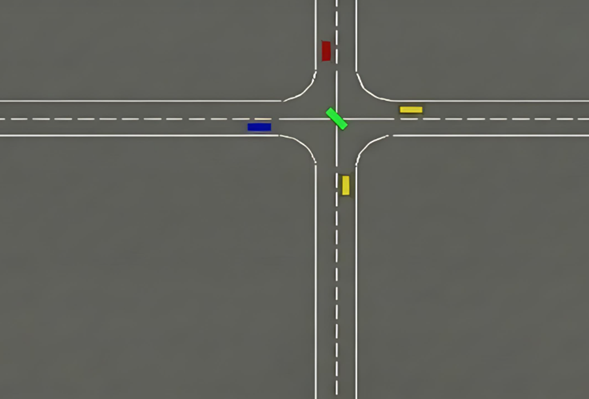}
        \caption{
        Conflict zone: vehicles converge at the intersection center, requiring coordinated timing to avoid collisions.
        }
        \label{fig:scenario_b}
    \end{subfigure}
    \caption{Example scenarios from the \emph{HighwayEnv} environment, illustrating initial vehicle placement and the resulting high-density conflict zone.}
    \label{fig:scenarios}
\end{figure}

At the start of each episode, five vehicles are positioned along their respective approach lanes at distances ranging from 0 to 75\,m from the intersection entry point, all initialized at 20\,m/s. Each vehicle's approach direction and intended trajectory are specified by the scenario definition. Of the five vehicles, three are controlled by MAPS learning agents, and two follow constant-speed trajectories (20\,m/s) to simulate non-cooperative background traffic. Assignments are fixed per scenario. Figure~\ref{fig:scenarios} illustrates an example initial configuration and the resulting conflict zone.

\subsubsection{Training and Evaluation Protocol}
Training is conducted over 900 episodes using the alternating optimization procedure of Algorithm~\ref{alg:joint-training}, with scenarios sampled uniformly at random from the 72 configurations. Evaluation is performed deterministically, removing all exploration noise across 100 episodes drawn uniformly from the same scenario distribution. Deterministic evaluation tests whether the learned policy has converged to a robust coordination strategy, and uniform sampling ensures coverage across the full range of traffic patterns. All methods use a fixed random seed for deterministic reproducibility; we discuss implications in Section~\ref{subsec:limitations}.

\subsection{Baseline Methods}
\label{subsec:baselines}

Table~\ref{tab:hyperparameters} (see Appendix) summarizes the training configuration used by our approach in all experiments. We compare MAPS against two recent methods representing distinct paradigms for multi-agent intersection coordination: value decomposition with demonstrations and actor-critic with adaptive exploration.

\paragraph{QMIXwD~\cite{Huang2023Multiagent}.} This baseline addresses exploration challenges by integrating learning from demonstrations into the QMIX value-decomposition framework. A pre-training stage leverages both expert demonstrations and self-generated interaction data to mitigate distributional shift. Following the original methodology, our implementation employs a loss combining a supervised margin term, a $TD(\lambda)$ loss, and $L_2$ regularization. The demonstration dataset consists of 10\% expert trajectories from a pre-trained greedy policy and 90\% self-generated data.

\paragraph{VN-MADDPG~\cite{Zhang2024VNMADDPGAV}.} This baseline extends the MADDPG actor-critic framework with variable-noise exploration and importance-sampling mechanisms to improve learning efficiency in continuous multi-agent action spaces.

\paragraph{Fair comparison.} All methods were implemented following the original papers, received identical observations, and were trained for the same 900 episodes on the same scenario distribution. Baseline hyperparameters follow the original publications.

\paragraph{Baseline selection rationale.} We selected baselines representing two dominant paradigms: value decomposition with demonstrations (QMIXwD) and actor-critic with adaptive exploration (VN-MADDPG). Other hierarchical methods from Section~2, such as MA-GA-DDPG~\cite{Liu2025Cooperative} and SafeR-ADAIM~\cite{Zhao2024CentralizedCF}, were not included because they rely on additional privileged mechanisms—level-$k$ game-theoretic priors and safety inspector modules, or risk-aware constrained optimization with domain-specific safety layers—that are external to the core learning architecture. Since our central claim concerns the \emph{proto-plan} coordination mechanism itself, we compare against methods that, like MAPS, rely solely on learned coordination from kinematic observations, ensuring performance differences are attributable to architectural design rather than supplementary safety modules.

\subsection{Evaluation Metrics}
\label{subsec:metrics}

We assess performance using four complementary metrics capturing safety, efficiency, and learning stability:

\begin{itemize}
    \item \textbf{Success Rate ($SR$):} The percentage of episodes in which all vehicles cross the intersection without collision: $SR = N_{\text{success}} / N_{\text{total}} \times 100\%$. The complementary collision rate is $CR = 100\% - SR$. Episodes reaching the 50-step time limit without all vehicles completing traversal are counted as failures.

    \item \textbf{Training Collision Count ($CC$):} The total number of collision events across all training episodes, $CC = \sum_{k=1}^{K} c_k$ where $c_k \in \{0,1\}$. This tracks learning safety, where lower values indicate fewer dangerous experiences during training.

    \item \textbf{Cumulative Episode Reward ($R$):} The total reward accumulated over an episode, $R = \sum_{t=1}^{T} r_t$. Higher values indicate better combined safety and efficiency.

    \item \textbf{Average Travel Time ($ATT$):} The mean number of simulation steps for active learning agents to traverse the intersection: $ATT = \frac{1}{|\mathcal{W}_{\text{active}}|} \sum_{i=1}^{|\mathcal{W}_{\text{active}}|} \tau_i$, where $\tau_i$ is the travel time for vehicle $i$. This reflects traversal efficiency: inefficient coordination leads to longer episodes, while early collisions artificially reduce step counts.
\end{itemize}


\section{Evaluation Results}
\label{sec:evaluation}

\subsection{Main Results}
\label{sec:main_results}

\begin{figure}[t]
    \centering
    \includegraphics[width=1.0\linewidth]{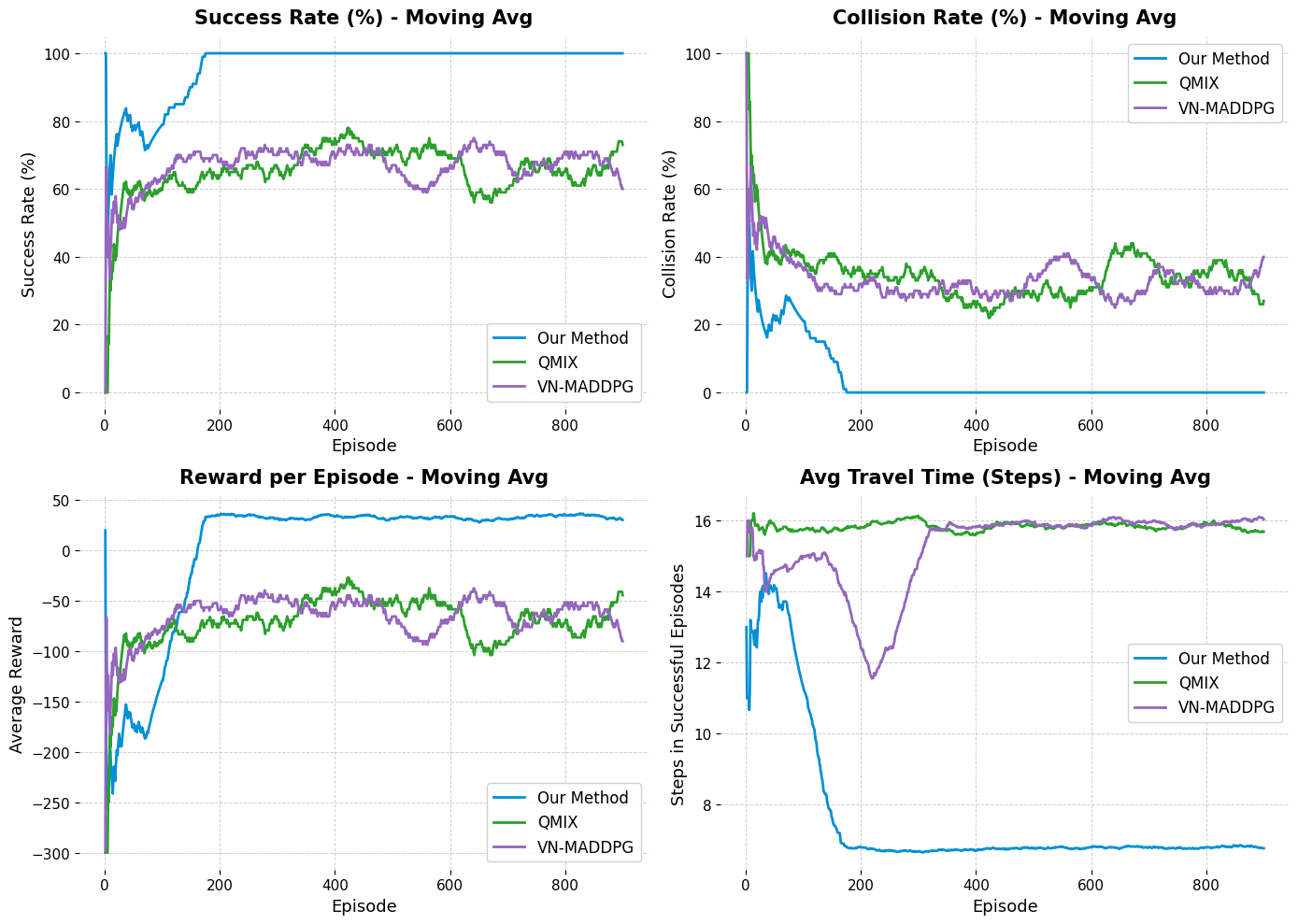}
    \caption{Training dynamics comparing MAPS and baselines: (a) success rate, (b) collision rate, (c) cumulative episode reward, and (d) average travel time. Moving average window size: 20.}
    \label{fig:results}
\end{figure}

Table~\ref{tab:crash_comparison} summarizes the comparative results, and Figure~\ref{fig:results} illustrates the learning dynamics during training. MAPS is the only approach to achieve collision-free evaluation while recording the lowest average travel time.

\begin{table}[t]
\centering
\caption{Comparison of collision counts and efficiency across methods.}
\label{tab:crash_comparison}
\renewcommand{\arraystretch}{1.2}
\setlength{\tabcolsep}{0pt}
\begin{tabular*}{\columnwidth}{@{\extracolsep{\fill}} l c c c }
\toprule
\textbf{Approach} & 
\begin{tabular}[c]{@{}c@{}}\textbf{Collisions} \\ (Train - 900)\end{tabular} & 
\begin{tabular}[c]{@{}c@{}}\textbf{Collisions} \\ (Eval - 100)\end{tabular} & 
\begin{tabular}[c]{@{}c@{}}\textbf{Avg Steps} \\ (Evaluation) \end{tabular} \\ 
\midrule
MAPS (ours)              & 21  & 0 & 7.8 \\
VN-MADDPG               & 132  & 35  & 12.7 \\
QMIXwD                  & 147  & 31  & 13.2 \\
\bottomrule
\end{tabular*}
\end{table}

\subsubsection{Safety Performance}
MAPS achieves \textbf{zero collisions} during evaluation (100\% success rate), compared to 35 collisions (65\% SR) for VN-MADDPG and 31 collisions (69\% SR) for QMIXwD.

The safety advantage extends to training. MAPS incurs only 21 collisions during 900 training episodes, an \textbf{84--85\% reduction} relative to VN-MADDPG (132) and QMIXwD (147), indicating that the hierarchical architecture learns safer behaviors more rapidly. A two-proportion z-test confirms these reductions are statistically significant ($p < 0.001$) for both baselines.

\subsubsection{Efficiency Performance}
MAPS demonstrates superior traversal efficiency: vehicles require an average of \textbf{7.8 simulation steps} to cross the intersection, compared to 12.7 for VN-MADDPG and 13.2 for QMIXwD, a \textbf{38\% reduction} over the best baseline. This gain stems from MAPS learning proactive coordination strategies that minimize unnecessary yielding, with the \emph{proto-plan} mechanism enabling Workers to anticipate coordination requirements rather than react to imminent conflicts.

\subsubsection{Learning Dynamics}
Figure~\ref{fig:results} reveals several notable patterns. MAPS converges to near-optimal performance within approximately 180 episodes, after which its learning curves exhibit markedly lower variance than either baseline. VN-MADDPG and QMIXwD show persistent oscillations in both success rate and reward throughout training, suggesting difficulty maintaining consistent coordination. Despite sharing the same 900-episode training budget, MAPS achieves substantially better final outcomes, indicating that the hierarchical decomposition yields a more tractable learning problem than flat multi-agent architectures.

\subsection{Ablation Studies}
\label{subsec:ablations}

\subsubsection{Contribution of the Master Agent}
\label{subsubsec:proto_ablation}
To isolate the contribution of the learned \emph{proto-plan}, we hold Worker policies fixed and manipulate only the Master's output during inference. We evaluate three conditions:
\begin{align}
z_{t}^{\text{trained}} &= \pi_{M}(s_{t}^{M}), \label{eq:proto_trained} \\
z_{t}^{\text{random}} &\sim \mathcal{U}(-1,1)^{d}, \label{eq:proto_random} \\
z_{t}^{\text{zero}} &= \mathbf{0}_{d}, \label{eq:proto_zero}
\end{align}
where $\pi_{M}$ is the trained Master policy. Under all conditions, Workers use the same trained policy $\pi_{W}$, with $s_{t}^{W_i}$ constructed by concatenating kinematic observations with the manipulated $z_t$.

\begin{figure}[!t]
    \centering
    \includegraphics[width=\linewidth]{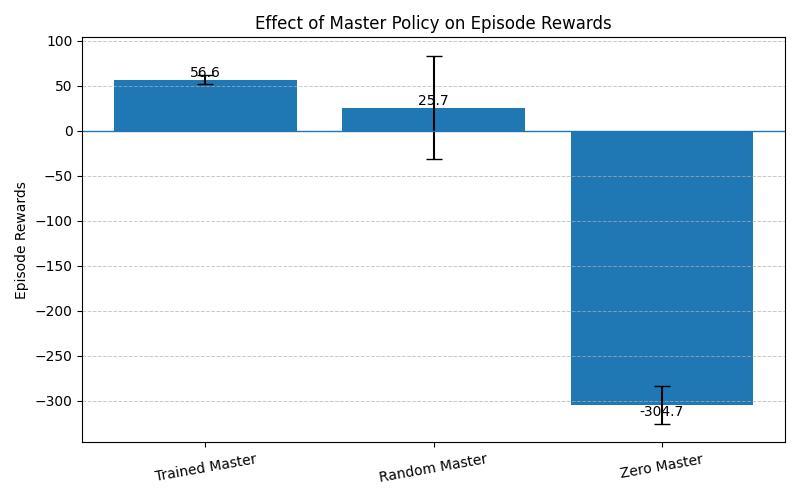}
    \caption{Cumulative episode reward under three \emph{proto-plan} conditions with Workers fixed. The trained Master significantly outperforms random and zero baselines, confirming that learned proto-plans encode essential coordination information.}
    \label{fig:master_reward_ablation}
\end{figure}

Figure~\ref{fig:master_reward_ablation} reports cumulative episode rewards over 400 evaluation episodes. The trained Master achieves the highest returns (mean reward: 56.6), random \emph{proto-plans} yield intermediate performance (25.7), and the zero-output condition collapses to strongly negative rewards ($-304.7$), a difference of over 360 reward points. These results demonstrate that the \emph{proto-plan} channel is essential for coordination and that Workers have learned to condition their behavior on its content.

\subsubsection{\emph{Proto-Plan} Embedding Analysis}
\label{subsubsec:proto_analysis}
\begin{figure}[!t]
    \centering
    \includegraphics[width=\linewidth]{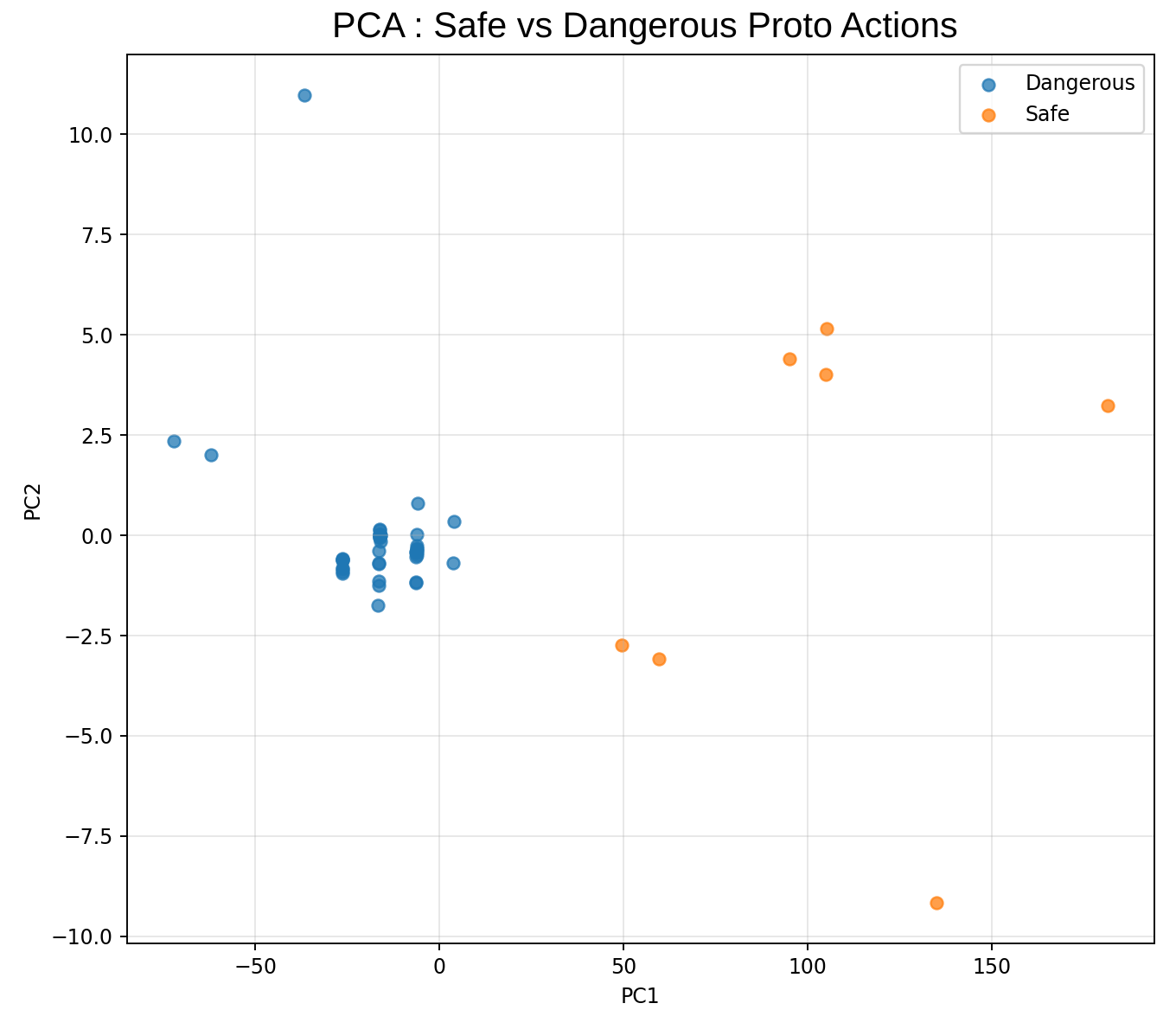}
    \caption{PCA projection of \emph{proto-plan} vectors, colored by safety predicate (dangerous: any inter-vehicle distance $<5$\,m). \emph{Proto-plans} encode safety-relevant context that Workers use for coordination decisions.}
    \label{fig:pca_master_proto}
\end{figure}

To interpret the information encoded by the Master, we project the $d$-dimensional \emph{proto-plan} vectors collected during evaluation to two dimensions using PCA and label each timestep by a safety predicate (dangerous if any inter-vehicle distance falls below 5\,m).
Figure~\ref{fig:pca_master_proto} reveals that \emph{proto-plans} occupy \textbf{distinct regions} for safe versus dangerous states, indicating that the Master has learned to compress global interaction risk into a compact representation. The first two principal components capture 76\% of the total variance (PC1: 63\%, PC2: 13\%).

\subsubsection{Sensitivity to Embedding Dimension}
\label{subsubsec:embedding_dim}
\begin{figure}[!t]
    \centering
    \includegraphics[width=\linewidth]{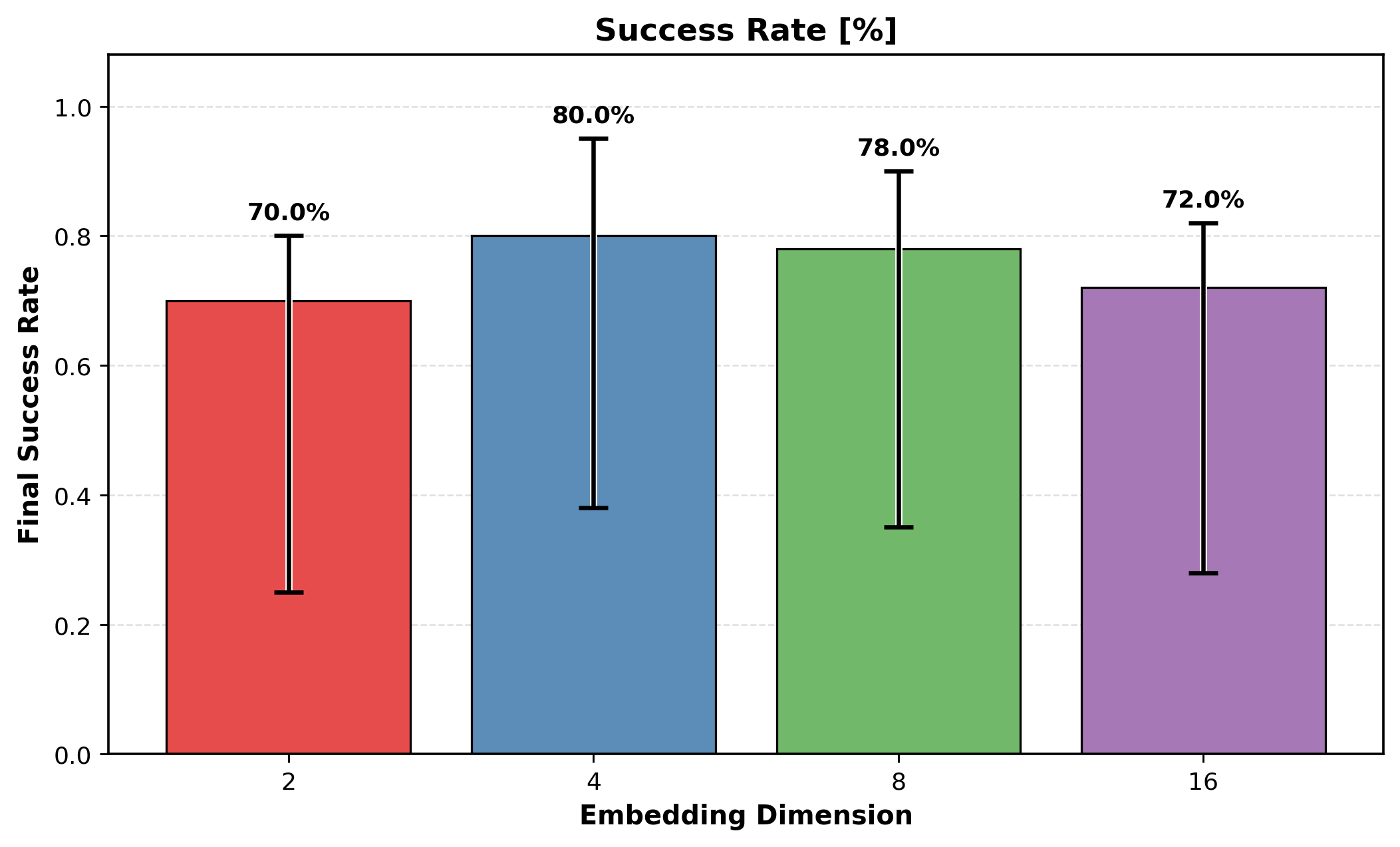}
    \caption{Impact of \emph{proto-plan} embedding dimension on success rate. Performance peaks at $d=4$ (80.0\%), indicating that a compact 4-dimensional vector provides the optimal balance between representational capacity and optimization stability.}
    \label{fig:embedding_dimension_success_rate}
\end{figure}

We evaluated the architecture across embedding dimensions $d \in \{2, 4, 8, 16\}$. As shown in Figure~\ref{fig:embedding_dimension_success_rate}, $d=4$ yields the best performance (80.0\% SR), compared to 70.0\% for $d=2$, 78.0\% for $d=8$, and 72.0\% for $d=16$.

\textbf{Note on evaluation protocols.} The success rates reported in this ablation study differ from those in Table~\ref{tab:crash_comparison} because the two experiments use different training configurations. Figure 6 uses a single-phase training setup to isolate the effect of embedding dimension, whereas Table 2 reports the final incrementally trained model (Phase 1 → Phase 2).

\begin{figure}[!t]
    \centering
    \includegraphics[width=\linewidth]{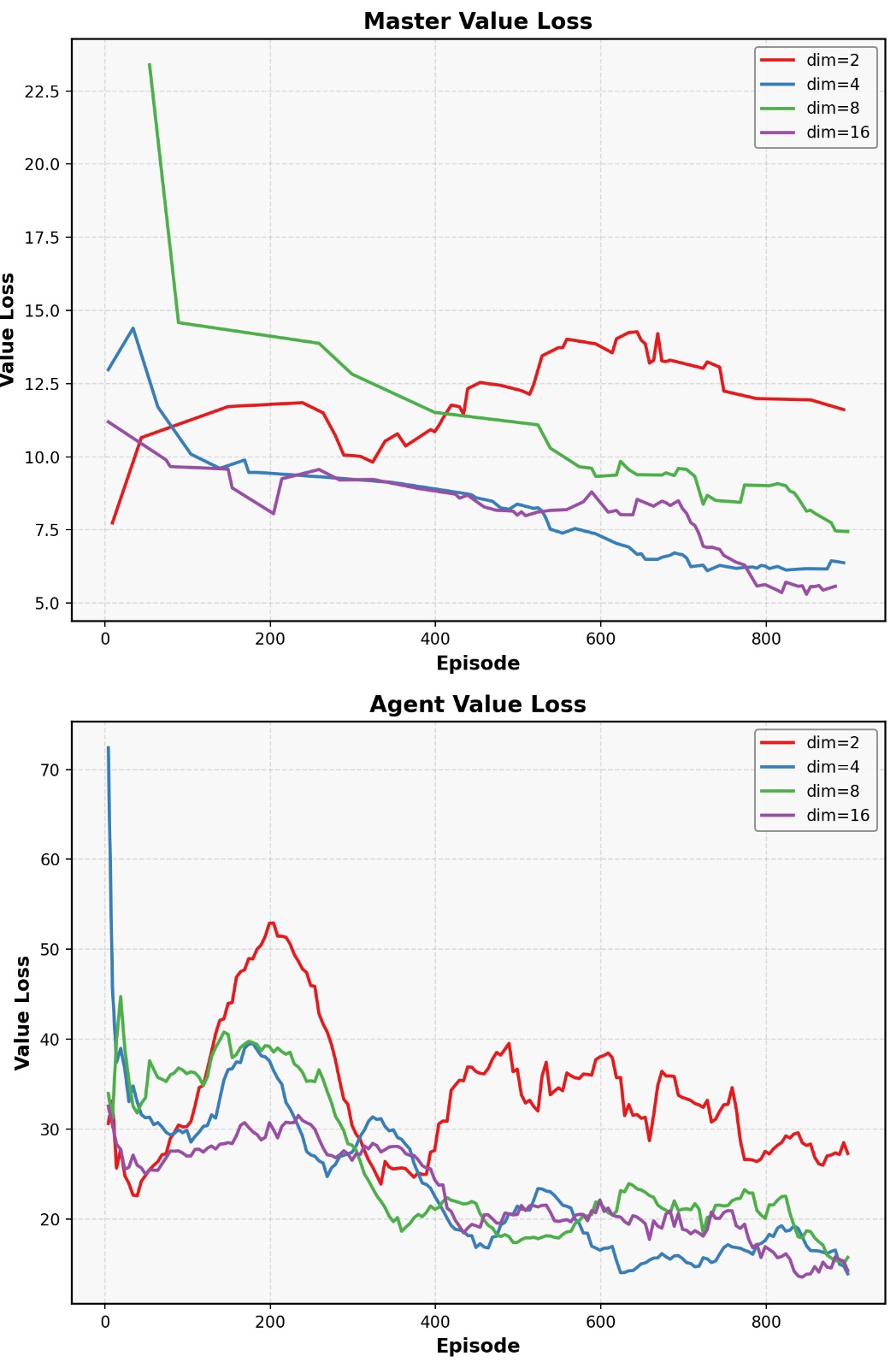}
    \caption{Master and Worker value-loss trajectories across embedding dimensions. The $d=4$ setting shows the most balanced and stable joint convergence across modules, with low final losses and reduced late-stage oscillations.}
    \label{fig:embedding_dimension_loss}
\end{figure}

\begin{figure}[!t]
    \centering
    \includegraphics[width=\linewidth]{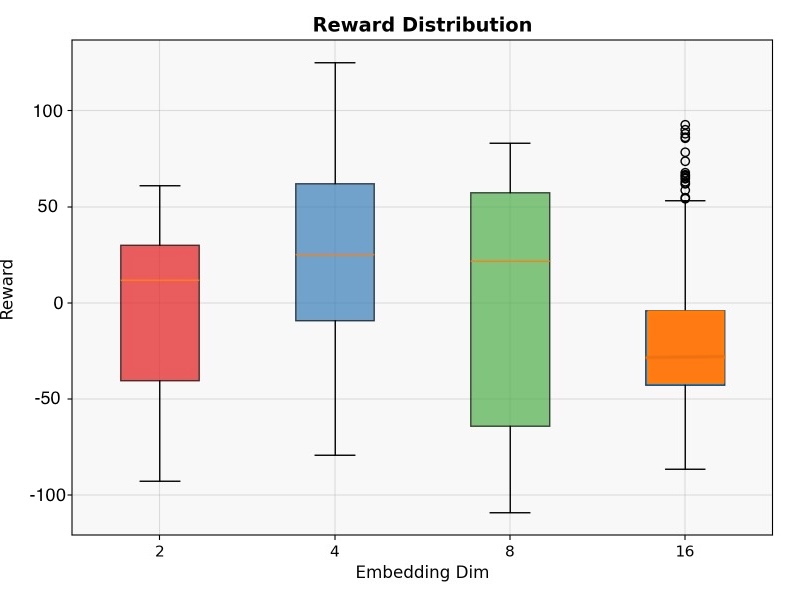}
    \caption{Reward distribution by embedding dimension. The $d=4$ policy exhibits the strongest central reward profile and robust dispersion, while higher-dimensional settings, especially $d=16$, show degraded central tendency and increased outlier behavior.}
    \label{fig:reward_dist}
\end{figure}

Analysis of training dynamics confirms that $d=4$ yields the most stable reward distribution. Figures~\ref{fig:embedding_dimension_loss} and~\ref{fig:reward_dist} provide direct optimization and reward evidence. Figure~\ref{fig:embedding_dimension_loss} shows that $d=4$ achieves the most balanced convergence across both Master and Worker value losses, with low final losses and reduced late-stage oscillations. Figure~\ref{fig:reward_dist} shows that $d=4$ also maximizes reward quality with the strongest central tendency and robust dispersion, while higher-dimensional settings (especially $d=16$) exhibit degraded central tendency and frequent outliers. Together, these results indicate that a 4-dimensional \emph{proto-plan} is the optimal operating point: it is stable to optimize, converges reliably to low-loss solutions, and yields the strongest policy-level returns.

This pattern reflects a trade-off between representational capacity and learnability: $d=2$ provides insufficient capacity to encode necessary coordination context, while $d \geq 8$ introduces optimization difficulty, leading to noisier policies. The result supports the central thesis that effective fleet coordination can be compressed into a very low-dimensional signal, making the \emph{proto-plan} mechanism practical even under bandwidth-constrained conditions.

\subsection{Zero-Shot Transfer to Unseen Fleet Sizes}
\label{subsec:generalization}

To evaluate whether the learned \emph{proto-plans} capture transferable coordination strategies, we test the model trained with three active agents (Section~\ref{subsec:incremental}) on five-agent configurations without fine-tuning. Across 100 test episodes, the system achieves a \textbf{94\% success rate}, demonstrating robust generalization to larger fleet sizes. The 6 failed episodes result from collisions in high-density or turning scenarios, with no significant pattern.

This result can be attributed to three architectural properties: (1) the \emph{proto-plan} embedding conveys coordination information in a format agnostic to the specific number of agents; (2) Worker parameter sharing ensures that control policies generalize across vehicle instances; and (3) the maximin reward structure encourages strategies that do not depend on particular agent configurations. Together, these properties indicate that \emph{proto-plan}-based coordination generalizes beyond the training regime without requiring additional learning.

\subsection{Limitations and Design Scope}
\label{subsec:limitations}

We identify several limitations that reflect deliberate scoping decisions and directions for future work.

\paragraph{Simulation fidelity.} Our evaluation uses HighwayEnv, a kinematic simulator that abstracts away perception noise, sensor latency, and detailed vehicle dynamics. This choice is intentional: it isolates the multi-agent coordination mechanism from confounding factors, providing a controlled testbed for validating the \emph{proto-plan} architecture. However, validation in higher-fidelity environments (e.g., SUMO or CARLA) is needed to confirm that coordination benefits transfer when richer dynamics and sensor noise are present.

\paragraph{Action space.} The current Worker action space is limited to binary longitudinal speed control (\texttt{accelerate}/\texttt{decelerate}). While this is a simplification, the architecture is agnostic to the Worker action space: continuous acceleration, lateral maneuvers, or richer discrete action sets can be substituted by modifying only the Worker policy head, without changes to the Master or the \emph{proto-plan} mechanism. Evaluating such extensions is a priority for future work.

\paragraph{Fleet size.} While the architecture demonstrates zero-shot transfer from three to five agents, the current Master uses a fixed-size concatenated input ($4N_{\max}$ features), which imposes a hard maximum. Scaling to substantially larger fleets would benefit from permutation-invariant input processing, such as attention-based or set-based architectures for the Master, which is a natural extension that preserves the \emph{proto-plan} broadcast mechanism. The present study validates the core coordination mechanism; large-scale deployment remains an open question.

\paragraph{Baseline scope.} We compare against two representative flat MARL methods. Hierarchical baselines that incorporate privileged mechanisms (game priors, safety inspectors) were excluded to ensure that performance differences reflect the coordination architecture rather than supplementary modules (see Section~\ref{subsec:baselines} for details).

\section{Conclusions and Future Work}
We presented MAPS, a hierarchical DRL architecture for multi-vehicle coordination at unsignalized intersections. The architecture uses a continuous \emph{proto-plan} embedding generated by a centralized Master and consumed by decentralized Workers, decoupling strategic coordination from tactical control while avoiding combinatorial action space scaling. Evaluations across 72 intersection configurations showed collision-free navigation with 38\% reduced travel time versus the best baseline, and ablation studies confirmed that \emph{proto-plans} encode meaningful coordination information. The architecture achieved 94\% success in zero-shot generalization from three to five agents.

The key insight is that multi-vehicle coordination compresses into a 4-dimensional continuous signal, enabling deployment in communication-constrained and bandwidth-limited V2X systems. Future work includes: (1) extending \emph{proto-plans} to encode \emph{long-horizon} strategies, amortizing communication cost across multiple time steps; (2) exploring \emph{hierarchical scaling} through mid-level masters that relay refined plans from higher nodes to local Workers; and (3) investigating \emph{heterogeneous fleets} via vehicle-class-specific Worker modules, leveraging modularity to integrate new agent types without retraining.

\bibliographystyle{ACM-Reference-Format}
\bibliography{bib}

\clearpage

\appendix
\section{The Hyperparameters of Our Proposed Approach}

\noindent\begin{minipage}{\columnwidth}
\centering
\small
\captionof{table}{Training Hyperparameters}
\setlength{\tabcolsep}{4pt}
\begin{tabular}{@{}p{0.64\columnwidth}p{0.30\columnwidth}@{}}
    \toprule
    \textbf{Parameter} & \textbf{Value} \\
    \midrule
    Algorithm & PPO \\
    Optimizer & Adam \\
    Loss Function & MSE \\
    Discount Factor ($\gamma$) & 0.99 \\
    Learning Rate (Master \& Agent) & 0.005 \\
    Batch Size & 32 \\
    Rollout Buffer Size ($N\_STEPS$) & 64 \\
    Clip Range (PPO) & 0.2 \\
    \midrule
    \multicolumn{2}{c}{\textbf{Additional Configuration}} \\
    \midrule
    Episodes Per Cycle & 300 \\
    Total Cycles & 3 \\
    Gradient Update Frequency & Every 2 Episodes \\
    Max Episode Time & 50 Steps \\
    Agent State Input Size & 8 \\
    \emph{proto-plan} Embedding Dimension & 4 \\
    Action Space Size & 2 \\
    Number of Agents & 5 \\
    Seed & 42 \\
    Deep learning framework & PyTorch \\
    RL library & Stable-Baselines3 \\
    Hardware & CPU \\
    Approximate training time & 20m \\
    \midrule
    \multicolumn{2}{c}{\textbf{Reward Structure}} \\
    \midrule
    Target Reached Reward & 50 \\
    Collision Reward & -300 \\
    Starvation Reward & -5 \\
    \bottomrule
    \label{tab:hyperparameters}
\end{tabular}
\end{minipage}
\par\medskip

\end{document}